%% file: Swap.tex
\documentclass[conference,a4paper]{IEEEtran}

\usepackage{CJKutf8}
\usepackage[switch]{lineno}
\usepackage{makecell}
\usepackage{graphicx}
\usepackage{upgreek}
\usepackage{CJK}
\usepackage{bbm}
\usepackage{hyperref} 
\usepackage{graphicx}
\usepackage{float}
\usepackage{lettrine}
\usepackage{booktabs}
\usepackage{bm}
\usepackage{amsmath}
\DeclareRobustCommand*{\IEEEauthorrefmark}[1]{\raisebox{0pt}[0pt][0pt]{\textsuperscript{\footnotesize #1}}}

\usepackage{algorithm}
\usepackage{algpseudocode}
\usepackage{fontawesome}
\usepackage[table,xcdraw]{xcolor}
\usepackage{soul}
\usepackage{amstext}
\usepackage{booktabs}
\usepackage{graphicx}
\usepackage{adjustbox}
\usepackage{float}
\usepackage{amsmath}
\usepackage{color}
\usepackage{multirow}

\begin{document}
\IEEEoverridecommandlockouts
\pagenumbering{roman}

%
    \title{Optimized Vectorizing of Building Structures with Switch: High-Efficiency Convolutional Channel-Switch Hybridization Strategy }

\definecolor{Moule}{rgb}{1.0,0.1,0.1}
\newcommand{\com}[1]{\textcolor{red}{\emph{#1}}}
\newcommand{\Moule}[1]{{\textbf{\color{Moule} #1}}}
\newcommand{\aka}{\emph{a.k.a.}}
\newcommand{\rref}[1]{\textcolor{red}{~\ref{#1}}}
\def\eg{\emph{e.g.}} \def\Eg{\emph{E.g}}
\def\ie{\emph{i.e.}} \def\Ie{\emph{I.e.}}
\def\cf{\emph{c.f.}} \def\Cf{\emph{C.f.}}
\def\etc{\emph{etc}} \def\vs{\emph{vs.}}
\def\aka{\emph{a.k.a.}}
\def\wrt{w.r.t.} \def\dof{d.o.f.}
\def\etal{\emph{et al.}}
\author{\IEEEauthorblockN{
Moule Lin$^{\href{https://orcid.org/0000-0001-6227-2392}{\includegraphics[scale=0.07]{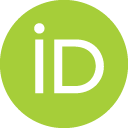}}}$,\thanks{The work described in this paper is supported by National Natural Science Foundation of China (32171777), Fundamental Research Funds for the Central Universities(2572017PZ04), Heilongjiang Province Applied Technology Research and Development Program Major Project(GA20A301) and China Scholarship Council. A. Jung was supported by project no. TKP2021-NVA-29, which has been implemented with the support provided by the Ministry of Culture and Innovation of Hungary from the National Research, Development and Innovation Fund, financed under the TKP2021-NVA funding scheme.}\thanks{(Corresponding author: Weipeng Jing.)}\thanks{M. Lin is with the College of Computer and Control Engineering, Northeast Forestry University, Harbin 150040, China. And with the Faculty of Informatics, Eötvös Loránd University, Budapest 1117, Hungary. (e-mail: linmoule@inf.elte.hu))}   
Weipeng Jing$^{\href{https://orcid.org/0000-0001-7933-6946}{\includegraphics[scale=0.07]{images/ID.png}}}$, \textit{Member, IEEE},\thanks{W. Jing and C. Li is with the College of Computer and Control Engineering, Northeast Forestry University, Harbin 150040, China.  (e-mail:  jwp@nefu.edu.cn; lichaonefuzyz@nefu.edu.cn)}   
Chao Li$^{\href{https://orcid.org/0000-0003-1932-7698}{\includegraphics[scale=0.07]{images/ID.png}}}$, \textit{ Member, IEEE} and
András Jung$^{\href{https://orcid.org/0000-0003-3250-4097}{\includegraphics[scale=0.07]{images/ID.png}}}$,\textit{ Member, IEEE}\thanks{A.Jung is with the Faculty of Informatics, Eötvös Loránd University, Budapest 1117, Hungary (e-mail: jung@inf.elte.hu).}  
}    
}
\maketitle
\input{section/0-abstract}

\input{section/1-introduction}

\input{section/3-propose}

\input{section/4-experiments}
\input{section/5-conclusion}
\input{section/6-Acknowledgement}
\IEEEpeerreviewmaketitle
\vspace{7pt}
\bibliographystyle{IEEEtran}
\bibliography{REFERENCE}
\end{document}

%% file: section/0-abstract.tex
\begin{abstract}

The building planar graph reconstruction, \aka~footprint reconstruction, which lies in the domain of computer vision and geoinformatics, has been long afflicted with the challenge of redundant parameters in conventional convolutional models. Therefore, in this letter, we proposed an advanced and adaptive shift architecture, namely the "Switch" operator, which incorporates non-exponential growth parameters while retaining analogous functionalities to integrate local feature spatial information, resembling a high-dimensional convolution operation.
The "Switch" operator, cross-channel operation, architecture implements the XOR operation to alternately exchange adjacent or diagonal features, and then
blends alternating channels through a 1x1 convolution operation to consolidate information from different channels. 
The SwitchNN architecture, on the other hand, incorporates a group-based parameter-sharing mechanism inspired by the convolutional neural network process and thereby significantly reducing the number of parameters.
We validated our proposed approach through experiments on the SpaceNet corpus, a publicly available dataset annotated with 2,001 buildings across the cities of Los Angeles, Las Vegas, and Paris. Our results demonstrate the effectiveness of this innovative architecture in building planar graph reconstruction from 2D building images.
\end{abstract}
{\IEEEkeywords Planar Graph, Footprint, Switch, SpaceNet, Reconstruction }

%% file: section/1-introduction.tex
\section{INTRODUCTION}
\lettrine[lines=2]{H}{uman} vision has evolved over millions of years to be highly adept at comprehending images, enabling us to perceive the depth and structure of objects in our environment. When we view a building, for example, we can quickly discern the different components that make up its structure, detect the corners and edges that can define its shape, and infer the graph relationships between these components. This ability to perceive structures from images is a fundamental aspect of human vision, and it allows us to make sense of the world around us.

However, for conventional computer vision~\cite{guo2022attention} and geoinformatics~\cite{yordanov2021overview}, the task of identifying these structures from images remains a significant challenge. While these fields have made great strides in recent years, they still struggle to replicate the efficiency and accuracy of the human visual system. This is due in part to the fact that conventional methods are often limited by their reliance on hand-crafted features and rigid graph structures, which can lead to errors and inaccuracies in the resulting reconstructions.
\begin{figure}[t]
    \centering
    \includegraphics[width=0.5\textwidth]{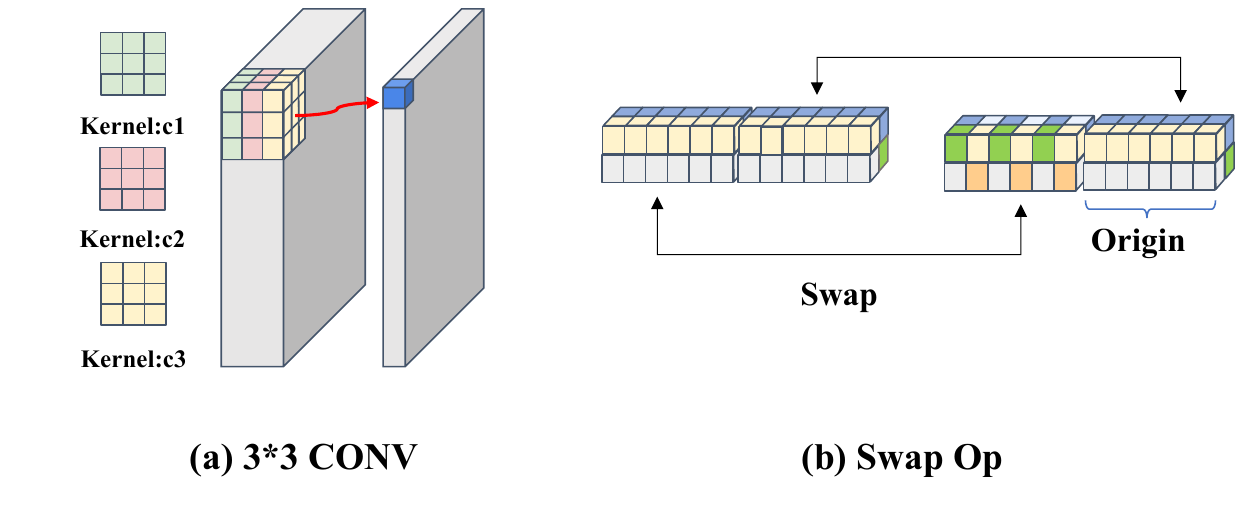}
    \caption{Schematic illustration of Convolutional Neural Network (CNN) and Switch: (a) 3*3 Conv operation; (b) Switch operation; The 3*3 convolutional operation and the Switch operation have similar functionality with integrating local spatial features, as stated in (a) and (b). However, the number of parameters in the convolutional operation can increase exponentially with the number of channels, while the Switch operation can effectively reduce the parameters and facilitate fine-tuning of the model.}
    \vspace{-0.5cm}
    \label{fig:difer}
\end{figure}
\begin{figure*}[t]
    \centering
    \includegraphics[width=1.0\textwidth]{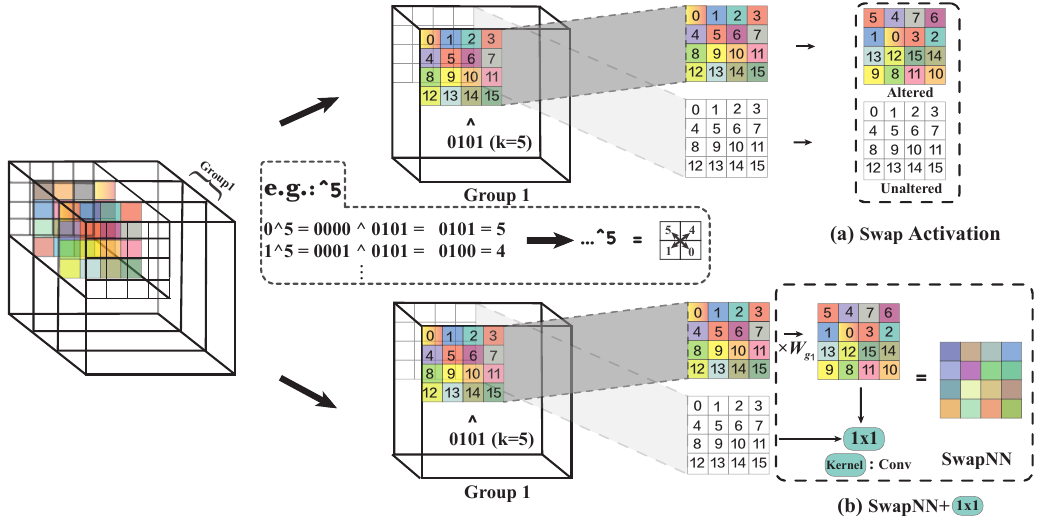}
   
    \caption{Schematic illustration of Switch and SwitchNN, k is xor parameter, taken k = 5 as an example in here. (a) Switch is an activation function, which only contains a Switch operation without any parameter. (b) SwitchNN constructed a parameter of W, shared in the group, and combined 1x1 convolution to mix the channel's features. Interval Switch adopted in adjacent channels, i.e. odd and even for channels index.}
    \vspace{-0.5cm}
    \label{fig:Switch}
\end{figure*}
\par
Building planar graph structures is an essential step towards creating digital representations that can be used for various purposes, such as visual representation~\cite{ewenstein2009knowledge}, analysis~\cite{handana2018performance}, and simulation~\cite{del2016large}. By converting physical structures into digital vector models, we can capture their geometric information and use it to study and understand the built environment. Vector models have become indispensable tools for architects, urban planners, and other professionals who deal with the built environment. They allow us to create accurate and detailed representations of buildings to explore various design options~\cite{boyko2012benchmarking}.
\par

\par
Nauata \etal~\cite{nauata2020vectorizing} introduced an algorithm that involves using Convolutional Neural Networks (CNNs) to identify geometric primitives and understand their relationships, along with Integer Programming (IP) to combine this information and create a 2D planar graph. 
It adopted CNNs to detect geometric primitives and their relationships can be computationally expensive and may require a significant amount of training data to achieve accurate results. Additionally, the algorithm's reliance on Integer Programming (IP) may result in slower execution times, as IP is known to be a computationally intensive optimization technique.
%
%
Zhang \etal~\cite{zhang2020conv} proposed a novel message-passing neural (MPN) architecture that employs convolutions to encode messages, thus effectively mitigating the problem of similar line features. 
\par
However, topology-based models often require an exhaustive search of all subspaces, which demands enormous computational resources. Conv-MPN requires at least two NVIDIA TitanX GPUs (24G RAM each) for model inference. 
%
%
Therefore, we will blaze a new pathway against those challenges by harnessing
flexible shift architecture with non-exponential growth parameters implemented by CUDA/C++ and Pytorch, superseding the traditional module of CNNs, to extract the
building boundary vectors efficiently and automatically. 
The main contributions of this research letter are summarized as follows:
1)To address aforementioned challenges, we introduce two advanced and adaptive shift architectures, Switch and SwitchNN implemented by CUDA/C++ and combined with Pytorch, which employ non-exponential growth parameters. 2)We validated our proposed approach through experiments on the SpaceNet corpus and it demonstrates the SOTA performance in the extraction of building boundary structures doamin.
\par
%


%% file: section/3-propose.tex
\section{Method}
\label{method}
Switch operator facilitates the efficient communication and exchanges~\cite{guo2022differentiable} of information between neighboring features. This approach allows each pixel's features to be blended with those of its neighbors, leading to enhanced accuracy in the reconstruction task.
\par
And then, Switch incorporates a 1x1 convolutional layer to mix the channel information of each pixel's local neighborhood. By doing so, it is able to capture more complex patterns and relationships between neighboring pixels. As shown in the Fig.~\ref{fig:Switch}.
It has been shown to yield comparable performance as the k $*$ k (k $>$ 1) convolutional operation while drastically reducing the number of parameters required. Specifically, the parameter count is equivalent to that of a 1x1 convolutional operation due to the absence of parameters in the Switch operation itself. Notably, this approach has proven to be highly effective in the context of small datasets for final adjustment.
%


\subsubsection{Switch}
Inherent spatial correlations, inferences, and similarities exist between adjacent features. In order to effectively incorporate such relationships, we utilize the XOR operation to intermix the neighboring channel dimensions of the features. 
For each selected channels, each feature exchange with its neighboring feature, also can exchange with farther positions, that is controlled by hyper-parameter K, as shown in the Fig.~\ref{fig:Switch}.
Furthermore, we restrict the selected channels with a size of c/4 (where c represents the number of channels). The first c/2 channels remain unaltered, while the remaining c/2 channels undergo cross-exchange, i.e., one channel is exchanged with itself, while the other is exchanged with its adjacent channel. It is described as follows:

\begin{equation}
\left\{\begin{matrix}

  C_{i}&=&C_{i}  & if & i&<=&c/2 \\
  C_{i}&=&C_{i\oplus{k}}  & else & i &>& c/2

\end{matrix}\right.
\end{equation}
where i is index of channel and $C_{i}$ is the channel. $c$ represents the total channels. $k$ is a hyperparameter to indicate which neighbor to be selected to exchange with $C_{i}$, such as $k=5$, and the following exchange rules are shown:
\[
\begin{alignedat}{3}
0\oplus{5}&=\verb|0000^0101| &=\verb|0101| &= 5 \\
1\oplus{5}&=\verb|0001^0101| &=\verb|0100| &= 4 \\
5\oplus{5}&=\verb|0101^0101| &=\verb|0000| &= 0 \\
4\oplus{5}&=\verb|0100^0101| &=\verb|0001| &= 1 \\
& \phantom{{} \oplus{5} 0000} \vdots
\end{alignedat}
\]
Here, cross-exchange is adopted due to $k=5$. This can also be regarded as a kernel size of 4, where different kernels generate varying exchange results, and the kernel size also will be changed.
\par
\subsubsection{SwitchNN}: SwitchNN is based on Switch and partitions channels into groups (\textbf{G} channels per group), with each group sharing a trainable parameter. It can be described as follows:
\begin{equation}
\left\{\begin{matrix}

  C_{i}&=&C_{i}*W_{g}  & if & i&<=&c/2 \\
  C_{i}&=&C_{i\oplus{k}}*W_{g}   & else & i &>& c/2
\end{matrix}\right.
\end{equation}
Where $W_{g}$ is a parameter shared with group g and others are the same with the Switch operation.
\par
Due to its parameterized nature, SwitchNN is more suitable for tasks involving large datasets, whereas Switch can be applied to tasks with smaller datasets, such as the building graph reconstruction, \aka~ footprint, in this letter.
SwitchNN presents a promising approach to channel reduction and feature exchange in convolutional neural networks, demonstrating the potential to achieve comparable performance to traditional convolutional operations while simultaneously reducing computational overhead.

\subsection{Edge Sampler}
\label{sampler}
For N corner points, there will be (N/(N-1))/2 candidate undirected edges generated, where N is the number of corners.
\begin{equation}
  E = (N/(N-1))/2
\end{equation}
E is the edges and N is the number of corners that we already know.
\par
All edges are treated as candidates, with two corners at each end. The output from U2Net passed the Sigmoid function criteria to get the score for each edge. And it can be split as following steps:
\par
1) Get a score, i.e. average value, along the edge generated by each pair corner.
\begin{equation}
  S_{0} = Avegave(B(C_{1},C_{2}))
\end{equation}
\par
2) Calculate its four, eight, and sixteen neighbors for each point(along the edge), and here, four neighbors were adopted, as shown in Fig.\ref{fig:four}.
\begin{equation}
  S_{1-4} = Avegave(B(C_{1_1\_4},C_{2_1\_4}))
\end{equation}
Here, a total of 5 scores are obtained, and the final score for the edges is computed as the average of these scores.

To meet the requirement of a fixed input shape for the final classification module, we adopted a strategy of expanding the candidate edges to 256 edges, which is to adjust the scores corresponding to each edge by subtracting the maximum score of the previous batch of edges.
\par
For instance, assuming there are 6 edges with scores of [0.1, 0.4, 0.6, 0.1, 0.2, 0.5] that need to be expanded to 16 edges, we calculated that the number of expansions required was 2, with a remainder of 4. During the first expansion, the corresponding scores were adjusted to [-0.5, -0.2, 0.0, -0.5, -0.4, -0.1]. The maximum score from the first expansion was then subtracted from the original scores, and the resulting scores were adjusted again for the second expansion. This process was repeated for the remainder as well. 
And it can be described as follows:
\begin{equation}
  S_{expansion} = S_{previous} - Max(S_{previous})
\end{equation}
\par
This approach ensured that each candidate edge has a score that is significant enough to be selected in the top-k edges during the subsequent selection process based on their scores. By incorporating this technique, we can expand the number of candidate edges while maintaining a balance between the available scores and the desired number of edges for the final classification module.
\par
In this letter, we adopted U2Net as common backbone and replaced all 3*3 Conv by Switch operator. 
\par
The U2Net consists of six output layers,  and only the features from the first four output layers are used in the sampler module, as the last two layers output features with a very small and meanless spatial resolution.
\par
Bilinear interpolation is used to extract features from these layers. 
\begin{footnotesize}
\begin{equation}
\begin{split}
  f(x,y) = f(Q_{11})(x_{2}-x)(y_{2}-y) + f(Q_{21})(x-x_{1})(y_{2}-y) \\ + f(Q_{12})(x_{2}-x)(y-y_{1}) + f(Q_{22})(x-x_{1})(y-y_{1}) 
\end{split}
\end{equation}
\end{footnotesize}
\noindent f(x, y) is pixel value in the feature map and $Q_{11},Q_{12},Q_{21},Q_{22}$ are its neighbor points, respestively. 
\par
And we have some rules to ensure the feature is the same for each U2Net output layer, to be more specific, for an input image with dimensions of 128x64x64, 64 points are sampled for each edge, resulting in a feature vector of length 64x128. For an input image with dimensions of 256x32x32, 32 points are sampled for each edge, resulting in a feature vector of length 32x256. The same shape of features is extracted from the first four output layers, and different weights are applied to each feature to combine them for the final feature.
\begin{equation}
  S_{i} = W_{i} * S(X_{i})
\end{equation}
S represents the feature, combined with four layers feature, for the input of the classification module. $W_{i}$ is the weight for layer {i}.
And a head used to generate final output that contains a single channel of 4x downsampled feature.

%% file: section/4-experiments.tex
\section{Experimentation}

\subsection{Data Detail}
\begin{table*}[t]
\centering
\caption{The table shows precision and recall values for our method, with the segmentation confidence threshold set to 0.8 and the classification confidence threshold set to 0.4. The best results are indicated in red, orange color indicate the MAE and Recall, respectively. And the FLOPs are calculated and demonstrate in the last column.}
\setlength{\tabcolsep}{2mm}\renewcommand{\arraystretch}{1.6}
\begin{tabular}{ccccccc}
\toprule[1pt]
\hline
 &
  \multicolumn{3}{c}{\textbf{Edges}} &
  \multicolumn{2}{c}{\textbf{Segmention}} &
  \multicolumn{1}{c}{\textbf{FLOPs}}
  \\ \cmidrule(lr){2-4}\cmidrule(lr){5-6}\cmidrule(lr){7-7}
\multirow{-2}{*}{\textbf{Models}} &
  \textbf{Preci. (\%)} &
  \textbf{F1(\%)} &
  \textbf{Time(ms)} &
  \textbf{MAE.} &
  \textbf{Recall(\%)} &
  \textbf{G}
  
  \\ \cmidrule(lr){2-4}\cmidrule(lr){5-6}\cmidrule(lr){7-7}
\textbf{PPGNet~\cite{zhang2019ppgnet}} &
  55.1 &
  52.8 &
  1.4 &
  {\color[HTML]{000000} \textbf{-}} &
  {\color[HTML]{000000} \textbf{-}}&
  {\color[HTML]{000000} \textbf{8.23}} \\
\textbf{LCNN~\cite{zhou2019end}} &
  51.0 &
  {\color[HTML]{000000} 59.4} &
  {\color[HTML]{000000} 4.1} &
  {\color[HTML]{000000} \textbf{-}} &
  {\color[HTML]{000000} \textbf{-}}&
  {\color[HTML]{000000} \textbf{36.31}} \\
\textbf{Conv-MPN~\cite{zhang2020conv}} &
  56.9 &
  58.7 &
  11.9 &
  {\color[HTML]{000000} \textbf{-}} &
  {\color[HTML]{000000} \textbf{-}}&
  {\color[HTML]{000000} \textbf{142.79}} \\
\textbf{Nauata~\etal~\cite{nauata2020vectorizing}} &
  {68.1} &
  56.3 &
  10.5 &
  {\color[HTML]{000000} \textbf{-}} &
  {\color[HTML]{000000} \textbf{-}}&
  {\color[HTML]{000000} \textbf{127.90}} \\ 
  \textbf{Zhang~\etal~\cite{zhang2021structured}} &
  {67.1} &
  62. &
  15.3 &
  {\color[HTML]{000000} \textbf{-}} &
  {\color[HTML]{000000} \textbf{-}}&
  {\color[HTML]{000000} \textbf{184.23}} \\ \textbf{HEAT~\cite{chen2022heat}} &
  {\color[HTML]{000000} 80.6} &
  76.2&
  7.3 &
  {\color[HTML]{000000} \textbf{-}} &
  {\color[HTML]{000000} \textbf{-}}&
  {\color[HTML]{000000} \textbf{56.8}} \\ \textbf{Roof-Former~\cite{zhao2024vectorizing}} &
  {\color[HTML]{000000} \textbf{82.3}} &
  \textbf{78.1}&
  9.7 &
  {\color[HTML]{000000} \textbf{-}} &
  {\color[HTML]{000000} \textbf{-}}&
  {\color[HTML]{000000} \textbf{84.41}} \\ 
  \hline
  
\textbf{Ours (Witout Switch)} &
  {\color[HTML]{FE0000} \textbf{81.2}} &
  {\color[HTML]{FE0000} \textbf{74.4}} &
  {\color[HTML]{FE0000} \textbf{8.4}} &
  {\color[HTML]{FE996B} \textbf{0.026}} &
  {\color[HTML]{FE996B} \textbf{79.81}}&
  {\color[HTML]{000000} \textbf{39.71}} \\
  \textbf{Ours (Switch + Conv 1*1)} &
  {\color[HTML]{FE0000} \textbf{82.9}} &
  {\color[HTML]{FE0000} \textbf{79.8}} &
  {\color[HTML]{FE0000} \textbf{7.2}} &
  {\color[HTML]{FE996B} \textbf{0.018}} &
  {\color[HTML]{FE996B} \textbf{83.7}}&
  {\color[HTML]{000000} \textbf{24.83}} \\
  \hline
  \bottomrule[1pt]
\end{tabular}
\label{result2}
\end{table*}
The initial data for this study was sourced from the SpaceNet corpus, a publicly available dataset hosted on Amazon Web Services (AWS). 
\par
This dataset contained annotations for 2,001 buildings across three cities: Los Angeles, Las Vegas, and Paris. The annotations were manually created by expert annotators and provided detailed information about the building's corners and edges. The use of this dataset allowed for the training and evaluation of the proposed building reconstruction approach.
The annotated data are from Fuyang \etal~\cite{nauata2020vectorizing}, containing corners and edges for all images.
\subsection{Experiments}
The Switch operator in Pytorch/C++ is an innovative approach that can significantly improve the efficiency and accuracy of machine learning algorithms and has integrated custom-built operators into Pytorch.
Hyperparameters, such as the learning rate, batch size, \etc. In this case, the learning rate has been set to 0.001, which is a common starting value for many machine learning models. 
The rate is decayed by 1e\-4 to prevent overfitting and improve the generalization of the model.
The batch size is 8.
One optimization that has been implemented is the rewriting of the sampler in Pytorch's DataLoader. By ensuring that the same corner images are sampled into the same batch, the GPU memory is better utilized, leading to improved training speed and reduced memory usage.
\par
A single NVIDIA Tesla V100 GPU with 32G memory is used for all the experiments and its memory usage is intensive due to the Switch operator. 
\par

The optimal recall value is attained by setting the segmentation threshold at a confidence level of 0.6, and the classification confidence threshold at 0.5 or 0.6. Conversely, the highest precision value is observed by setting the segmentation confidence threshold to 0.8, and the classification threshold to 0.7.
%

\par
\begin{figure*}[htb]
    \centering
    \includegraphics[width=0.9\textwidth]{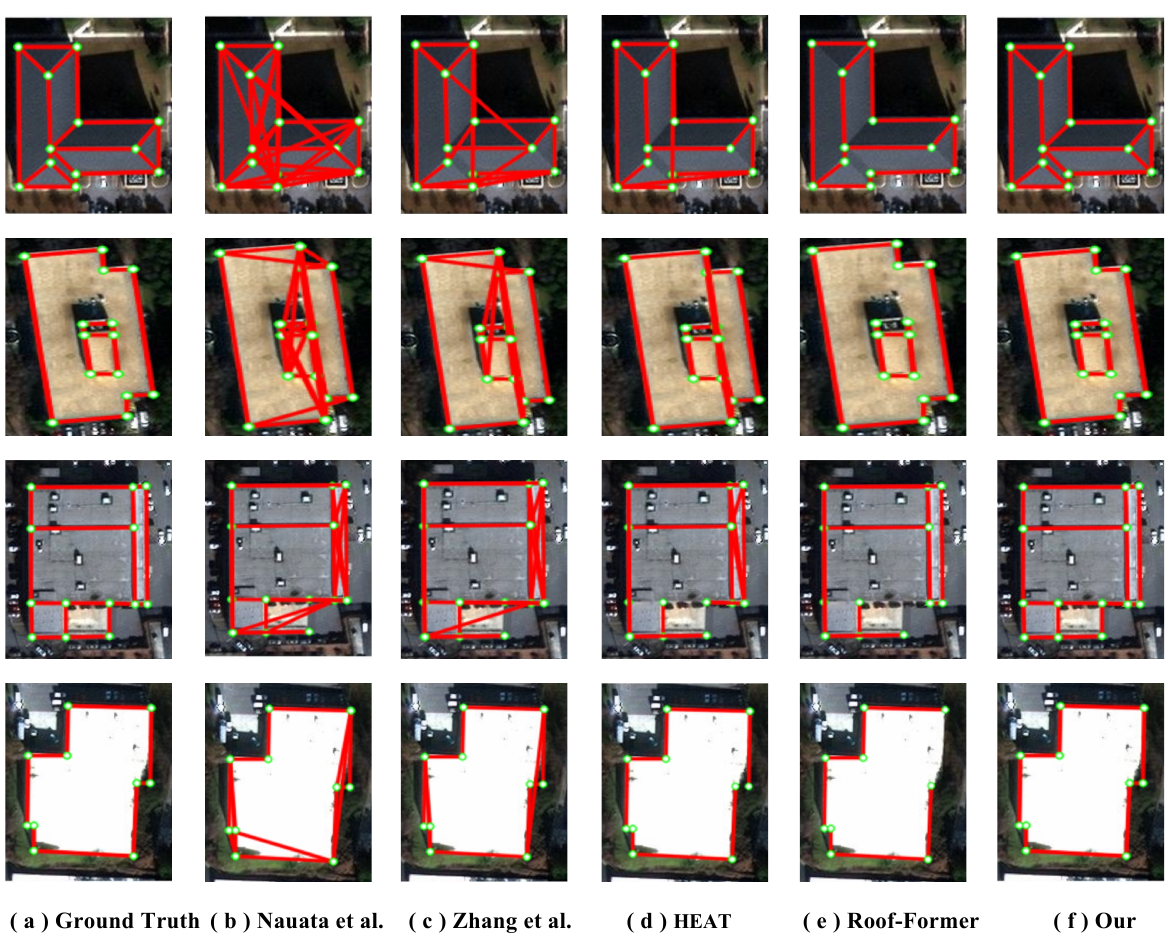}
    \caption{Building Planar Graph Reconstruction Results: (a) Ground Truth; (b) Nauata \etal; (c) Zhang \etal; (d) HEAT; (e) Roof-Former; (f) Our}
    \vspace{-0.5cm}
    \label{fig:result1}
\end{figure*}

Fig.\rref{fig:result1} and Table.\rref{result2} show representative building planar graph reconstruction results by our algorithm, and some comparative methods: PPGNet~\cite{zhang2019ppgnet}, LCNN~\cite{zhou2019end}, Conv\-MPN~\cite{zhang2020conv},  Nauata \etal~\cite{nauata2020vectorizing}, HEAT~\cite{chen2022heat} and Roof-Former~\cite{zhao2024vectorizing}. The table compares our method with existing approaches in terms of edge detection, segmentation, and computational efficiency. Our method achieves competitive precision (81.24\% without switch, 82.9\% with switch) and F1 score (74.41\% without switch, 79.8\% with switch), outperforming several prominent methods. Additionally, our method demonstrates efficient computation, with inference times of 8.4ms and 7.2ms respectively. Overall, our approach shows superior performance, highlighting its effectiveness in boundary vector extraction tasks.

%% file: section/5-conclusion.tex
\section{CONCLUSION}
This letter introduces an innovative algorithm for graph reconstruction in the context of building construction. 
In here, we proposed the Switch operator as a viable alternative to high filter size convolution operations, such as the 3 * 3 filter Conv operation. By leveraging its similar functionality, the Switch operation effectively reduces the number of parameters required. It is important to note that the Switch operator is not limited to the building graph reconstruction task alone; it has the potential for deployment in any deep learning domain that necessitates real-time responsiveness and mobile device compatibility.

%% file: section/6-Acknowledgement.tex
\section{Acknowledgement }
We would like to thank the Amazon Web Services (AWS) provided dataset of SpaceNet and thanks Nauata \etal annotated 2,001 buildings across Los Angeles, Las Vegas and Paris.

%% file: Swap.bbl
\begin{thebibliography}{10}
\providecommand{\url}[1]{#1}
\csname url@samestyle\endcsname
\providecommand{\newblock}{\relax}
\providecommand{\bibinfo}[2]{#2}
\providecommand{\BIBentrySTDinterwordspacing}{\spaceskip=0pt\relax}
\providecommand{\BIBentryALTinterwordstretchfactor}{4}
\providecommand{\BIBentryALTinterwordspacing}{\spaceskip=\fontdimen2\font plus
\BIBentryALTinterwordstretchfactor\fontdimen3\font minus \fontdimen4\font\relax}
\providecommand{\BIBforeignlanguage}[2]{{%
\expandafter\ifx\csname l@#1\endcsname\relax
\typeout{** WARNING: IEEEtran.bst: No hyphenation pattern has been}%
\typeout{** loaded for the language `#1'. Using the pattern for}%
\typeout{** the default language instead.}%
\else
\language=\csname l@#1\endcsname
\fi
#2}}
\providecommand{\BIBdecl}{\relax}
\BIBdecl

\bibitem{guo2022attention}
M.-H. Guo, T.-X. Xu, J.-J. Liu, Z.-N. Liu, P.-T. Jiang, T.-J. Mu, S.-H. Zhang, R.~R. Martin, M.-M. Cheng, and S.-M. Hu, ``Attention mechanisms in computer vision: A survey,'' \emph{Computational Visual Media}, vol.~8, no.~3, pp. 331--368, 2022.

\bibitem{yordanov2021overview}
V.~Yordanov, L.~Biagi, X.~Truong, V.~Tran, and M.~Brovelli, ``An overview of geoinformatics state-of-the-art techniques for landslide monitoring and mapping,'' \emph{The International Archives of Photogrammetry, Remote Sensing and Spatial Information Sciences}, vol.~46, pp. 205--212, 2021.

\bibitem{ewenstein2009knowledge}
B.~Ewenstein and J.~Whyte, ``Knowledge practices in design: the role of visual representations asepistemic objects','' \emph{Organization studies}, vol.~30, no.~1, pp. 07--30, 2009.

\bibitem{handana2018performance}
M.~Handana, R.~Karolina \emph{et~al.}, ``Performance evaluation of existing building structure with pushover analysis,'' in \emph{IOP Conference Series: Materials Science and Engineering}, vol. 309, no.~1.\hskip 1em plus 0.5em minus 0.4em\relax IOP Publishing, 2018, p. 012039.

\bibitem{del2016large}
M.~Del Carpio~Ramos, G.~Mosqueda, and M.~J. Hashemi, ``Large-scale hybrid simulation of a steel moment frame building structure through collapse,'' \emph{Journal of Structural Engineering}, vol. 142, no.~1, p. 04015086, 2016.

\bibitem{boyko2012benchmarking}
C.~T. Boyko, M.~R. Gaterell, A.~R. Barber, J.~Brown, J.~R. Bryson, D.~Butler, S.~Caputo, M.~Caserio, R.~Coles, R.~Cooper \emph{et~al.}, ``Benchmarking sustainability in cities: The role of indicators and future scenarios,'' \emph{Global Environmental Change}, vol.~22, no.~1, pp. 245--254, 2012.

\bibitem{nauata2020vectorizing}
N.~Nauata and Y.~Furukawa, ``Vectorizing world buildings: Planar graph reconstruction by primitive detection and relationship inference,'' in \emph{Computer Vision--ECCV 2020: 16th European Conference, Glasgow, UK, August 23--28, 2020, Proceedings, Part VIII 16}.\hskip 1em plus 0.5em minus 0.4em\relax Springer, 2020, pp. 711--726.

\bibitem{zhang2020conv}
F.~Zhang, N.~Nauata, and Y.~Furukawa, ``Conv-mpn: Convolutional message passing neural network for structured outdoor architecture reconstruction,'' in \emph{Proceedings of the IEEE/CVF Conference on Computer Vision and Pattern Recognition}, 2020, pp. 2798--2807.

\bibitem{guo2022differentiable}
S.~Guo, X.~Yang, J.~Ma, G.~Ren, and L.~Zhang, ``A differentiable two-stage alignment scheme for burst image reconstruction with large shift,'' in \emph{Proceedings of the IEEE/CVF Conference on Computer Vision and Pattern Recognition}, 2022, pp. 17\,472--17\,481.

\bibitem{zhang2019ppgnet}
Z.~Zhang, Z.~Li, N.~Bi, J.~Zheng, J.~Wang, K.~Huang, W.~Luo, Y.~Xu, and S.~Gao, ``Ppgnet: Learning point-pair graph for line segment detection,'' in \emph{Proceedings of the IEEE/CVF Conference on Computer Vision and Pattern Recognition}, 2019, pp. 7105--7114.

\bibitem{zhou2019end}
Y.~Zhou, H.~Qi, and Y.~Ma, ``End-to-end wireframe parsing,'' in \emph{Proceedings of the IEEE/CVF International Conference on Computer Vision}, 2019, pp. 962--971.

\bibitem{zhang2021structured}
F.~Zhang, X.~Xu, N.~Nauata, and Y.~Furukawa, ``Structured outdoor architecture reconstruction by exploration and classification,'' in \emph{Proceedings of the IEEE/CVF International Conference on Computer Vision}, 2021, pp. 12\,427--12\,435.

\bibitem{chen2022heat}
J.~Chen, Y.~Qian, and Y.~Furukawa, ``Heat: Holistic edge attention transformer for structured reconstruction,'' in \emph{Proceedings of the IEEE/CVF Conference on Computer Vision and Pattern Recognition}, 2022, pp. 3866--3875.

\bibitem{zhao2024vectorizing}
W.~Zhao, C.~Persello, X.~Lv, A.~Stein, and M.~Vergauwen, ``Vectorizing planar roof structure from very high resolution remote sensing images using transformers,'' \emph{International Journal of Digital Earth}, vol.~17, no.~1, pp. 1--15, 2024.

\end{thebibliography}
